# 基于图滤波与自表示的无监督特征选择算法

梁云辉[1,2], 甘舰文[1,2], 陈 艳[3], 周 芃[4], 杜 亮[1,2]
(1. 山西大学 计算机与信息技术学院, 太原 030006; 2. 山西大学 大数据科学与产业研究院, 太原 030006;
3. 四川大学 计算机学院, 成都 610065; 4. 安徽大学 计算机科学与技术学院, 合肥 230601)

**摘要:** 针对现有方法未考虑数据的高阶邻域信息而不能完全捕捉数据内在结构的问题, 提出一种基于图滤波与自表示的无监督特征选择算法. 首先, 将高阶图滤波器应用于数据获得其平滑表示, 并设计一个正则化器联合高阶图信息进行自表示矩阵学习以捕捉数据的内在结构; 其次, 应用 $l_{2,1}$ 范数重建误差项和特征选择矩阵, 以增强模型的鲁棒性与稀疏性选择判别的特征; 最后, 用一个迭代算法有效地求解所提出的目标函数, 并进行仿真实验以验证该算法的有效性.

# Unsupervised Feature Selection Algorithm Based on Graph Filtering and Self-representation

LIANG Yunhui[1,2], GAN Jianwen[1,2], CHEN Yan[3], ZHOU Peng[4], DU Liang[1,2]
(1. *School of Computer and Information Technology*, *Shanxi University*, *Taiyuan* 030006, *China*;
2. *Institute of Big Data Science and Industry*, *Shanxi University*, *Taiyuan* 030006, *China*;
3. *College of Computer*, *Sichuan University*, *Chengdu* 610065, *China*;
4. *School of Computer Science and Technology*, *Anhui University*, *Hefei* 230601, *China*)

**Abstract:** Aiming at the problem that existing methods could not fully capture the intrinsic structure of data without considering the higher-order neighborhood information of the data, we proposed an unsupervised feature selection algorithm based on graph filtering and self-representation. Firstly, a higher-order graph filter was applied to the data to obtain its smooth representation, and a regularizer was designed to combine the higher-order graph information for the self-representation matrix learning to capture the intrinsic structure of the data. Secondly, $l_{2,1}$ norm was used to reconstruct the error term and feature selection matrix to enhance the robustness and row sparsity of the model to select the discriminant features. Finally, an iterative algorithm was applied to effectively solve the proposed objective function and simulation experiments were carried out to verify the effectiveness of the proposed algorithm.

**Keywords:** graph filtering; self-representation; sparse; unsupervised feature selection

处理大量高维数据不但需要花费大量的计算资源, 而且包含许多噪声特征会降低算法的学习性能. 为选择最有价值的特征提升数据处理效率, 特征选择近年来备受关注, 并在文本分类[1]、图像处

**收稿日期:** 2023-05-04.
**第一作者简介:** 梁云辉(1998—), 男, 汉族, 博士研究生, 从事数据挖掘的研究, E-mail: 2350629530@qq.com. **通信作者简介:** 杜 亮(1985—), 男, 汉族, 博士, 副教授, 从事数据挖掘和机器学习的研究, E-mail: duliang@sxu.edu.cn.
**基金项目:** 国家自然科学基金面上项目(批准号: 61976129; 62176001; 62376146).

理[2]和生物信息学[3]等领域广泛应用. 根据是否使用标签信息，特征选择方法可分为有监督特征选择、半监督特征选择和无监督特征选择[4]. 根据特征选择策略的不同，特征选择方法又可分为过滤式方法、包装式方法和嵌入式方法[5]. 由于无法使用标签信息，无监督特征选择更具有挑战性，因此本文主要考虑无监督特征选择算法.

在无监督场景中，特征选择的关键是如何从原始特征空间中估计数据的内在结构，并识别准确保留结构的特征子集. 为估计潜在的结构信息，不同类型的图结构与数据的自表示学习得到广泛研究. Chen 等[6]将学习灵活最优图用于特征选择; Yuan 等[7]通过具有自适应图约束的凸非负矩阵分解进行选择特征; You 等[8]使用多组自适应图进行特征选择; Zhang 等[9]提出了一种基于自适应图学习和约束的无监督特征选择算法; Du 等[10]通过保持全局和局部结构，学习了用于特征选择的自适应图; Nie 等[11]结合 $l_{2,0}$范数约束与优化图进行特征选择; Zhu 等[12]选择了能很好重建原始数据的特征; Zhao 等[13]提出了一种带有数据重构的图正则化特征选择; Li 等[14]将重构函数学习过程嵌入到特征选择中.

尽管上述方法取得了很好的性能，但仍存在一定的局限性: 1) 数据不是平滑的，易受噪声异常值的影响; 2) 在建模时忽略了图结构的高阶邻域信息，不能实现更细致聚类结构的识别. 针对这些不足，本文提出一种基于图滤波与自表示的无监督特征选择算法. 首先构造 $k$ 近邻图，通过其不同阶邻域图的加权组合得到一个富含高阶邻域信息的图滤波器，并将图滤波器作用于数据得到其平滑表示，然后同时执行自表示学习和特征选择. 为更细致地捕捉数据的内在结构，设计一个正则化器以最小化自表示矩阵与图滤波器之间的差异，利用其高阶邻域信息实现更好的数据重建. 此外，将 $l_{2,1}$范数应用于重建误差项和特征选择项，以增强其鲁棒性与行稀疏性，实现更好的特征选择性能. 本文算法将图滤波器作用于原始数据获得其平滑表示，聚类假设相邻较近的点更可能位于同一簇中，因此平滑后数据有助于后续的特征选择任务; 将数据图结构的高阶邻域信息融入自表示学习中以实现对数据内在结构更好的重构.

# 1 预备知识

## 1.1 基本概念

对数据 $X \in \mathbb{R}^{n\times d}$，$n$ 表示样本数量，$d$ 表示特征维度，其中 $X$ 的第 $i$ 行表示为 $X_i \in \mathbb{R}^{1\times d}$，$l_{2,1}$范数定义为

$$\|X\|_{2,1} = \sum_{i=1}^{n} \sqrt{\sum_{j=1}^{d} X_{ij}^2}.$$

## 1.2 图滤波器

真实世界的信号通常是平滑的，图滤波器可在尽量保留图像细节特征的基础上对目标图像的噪声进行抑制，是图像处理中不可或缺的操作. 数据的图结构作为表示数据局部结构的一种重要方式在无监督特征选择任务中具有重要作用，本文采用的 $k$ 近邻图具体形式可表示为

$$S_{ij} = \begin{cases} \exp\left\{\dfrac{-\|X_i - X_j\|^2}{\delta^2}\right\}, & X_i \text{ 与 } X_j \text{ 是邻居}, \\ 0, & \text{其他}, \end{cases} \tag{1}$$

其中 $\delta$ 为高斯函数中带宽，其值为不同样本间欧氏距离的中值. 尽管直接使用相似性图 $S \in \mathbb{R}^{n\times n}$ 可获得良好的结果，但仍存在一定的局限性. 相似性图 $S$ 不是光滑的，图中元素大部分为 0. 相似性图仅考虑一阶邻域关系，这样的权重不能完全捕捉样本与簇之间的联系.

根据相似性图 $S$ 可得到其对应的对角矩阵 $D_S \in \mathbb{R}^{n\times n}$、对称转换矩阵 $G = D_S^{-1/2} S D_S^{-1/2}$ 和归一化 Laplace 矩阵 $L = I - G$，$D_S$ 中第 $i$ 个对角元素为 $\sum_{j=1}^{d} S_{ij}$. 样本之间的二阶关系可通过 $G^2 = G\times G$ 获得. 类似地，可使用一系列矩阵 $G^2, G^3, \cdots, G^\infty$ 获得样本的不同高阶邻域信息. 将这些矩阵以如下形式聚合获得图滤波器:

$$A = \sum_{t=0}^{\infty} w_t G^t, \tag{2}$$

其中 $w_t$ 为权重系数，$G^0$ 为单位矩阵. 采用热核扩散过程确定不同图聚合的权重系数，即

$$w_t = e^{-\eta} \frac{\eta^t}{t!}, \tag{3}$$

其中 $\eta$ 表示温度的非负值，且 $\sum_{t=0}^{\infty} w_t = 1, w_t \in [0,1]$. 通过合并式(2)与式(3)，可得图滤波器的具体形式为

$$A = e^{-\eta} \sum_{t=0}^{\infty} \frac{\eta^t}{t!} G^t = \sum_{t=0}^{\infty} \frac{(-\eta)^t}{t!} L^t = \exp\{-\eta L\}. \tag{4}$$

其中最后两个方程根据矩阵指数的定义成立. 基于此图滤波器可获得数据的平滑表示，有助于更细致地捕捉数据的内在结构以提升选择特征的质量.

### 1.3　自表示学习

自表示学习的核心思想是数据中每个样本都可以表示为其相关样本的线性组合. 如果一个样本很重要，则它将参与大多数其他样本的表示，从而产生一行重要的表示系数，反之亦然. 如果每个样本 $X_i$ 可通过其他样本的线性组合重建，则可得以下基于自表示的重建误差模型:

$$\min \sum_{i=1}^{n} \left\| X_i - \sum_{j=1}^{n} Z_{ij} X_j \right\|_2^2, \tag{5}$$

其中 $Z \in \mathbb{R}^{n\times n}$ 为自表示矩阵，用来度量第 $j$ 个样本对第 $i$ 个样本重构的贡献. 此外，为保证概率分布与非负性，本文对矩阵 $Z$ 施加约束 $Z\mathbf{1}_n = \mathbf{1}_n$，$Z \geqslant 0$，其中 $\mathbf{1}_n$ 为全为 1 的向量. $Z\mathbf{1}_n = \mathbf{1}_n$ 表示矩阵 $Z$ 中每行求和为 1 使得数据点位于仿射子空间的并集中，$Z \geqslant 0$ 使得自表示矩阵的自表示系数为非负值.

实际应用中的数据通常包含许多噪声，由于重建误差采用平方损失，上述模型对数据异常值非常敏感. 如果存在较大的偏差值，它将占据主导地位并严重降低模型的性能. 为提高模型(5)的鲁棒性，本文将其重写为

$$\min \sum_{i=1}^{n} d_i \left\| X_i - \sum_{j=1}^{n} Z_{ij} X_j \right\|_2, \tag{6}$$

其中 $d_i = \frac{1}{2} \left\| X_i - \sum_{j=1}^{n} Z_{ij} X_j \right\|_2$ 是衡量数据重建重要性的自适应权重. 当数据重建误差大时 $d_i$ 较小，当数据重建误差小时 $d_i$ 较大. 为简便，模型(6)可改写为矩阵形式:

$$\begin{aligned} &\min \| X - ZX \|_{2,1}, \\ &\text{s.t. } Z\mathbf{1}_n = \mathbf{1}_n, \quad Z \geqslant 0. \end{aligned} \tag{7}$$

对重建误差项采用 $l_{2,1}$ 范数可有效增强其鲁棒性以得到更可靠的自表示矩阵 $Z$.

## 2　算法框架

### 2.1　模型的建立

通过将多维数据点视为多个图信号，图滤波器可应用于数据矩阵 $X$:

$$\bar{X} = \exp\{-\eta L\} X, \tag{8}$$

得到的 $\bar{X}$ 为数据平滑表示，其有助于后续的聚类任务. 与此同时，在式(7)的基础上添加一个新的正则化器以进一步利用图滤波器 $A$ 的高阶邻域信息指导自表示矩阵 $Z$ 的学习:

$$\begin{aligned} &\min \| \bar{X} - Z\bar{X} \|_{2,1} + \alpha \| Z - A \|_F^2, \\ &\text{s.t. } Z\mathbf{1}_n = \mathbf{1}_n, \quad Z \geqslant 0. \end{aligned} \tag{9}$$

为同时执行特征选择任务与自表示学习，本文结合模型(9)得到基于图滤波与自表示的无监督特征选择算法(GFASR)的目标函数:

$$\begin{aligned}&\min \|\bar{X}W - Z\bar{X}W\|_{2,1} + \alpha\|Z - A\|_F^2 + \lambda\|W\|_{2,1},\\&\text{s.t. } W^TW = I,\quad Z\mathbf{1}_n = \mathbf{1}_n,\quad Z \geqslant 0,\end{aligned} \tag{10}$$

其中: $W \in \mathbb{R}^{d\times c}$是将高维数据投影到低维子空间的投影矩阵,也是用于特征选择的权重矩阵; $Z \in \mathbb{R}^{n\times n}$为自表示矩阵,用于测量理想邻居对重建每个样本的贡献; $A \in \mathbb{R}^{n\times n}$是根据相似性矩阵 $S \in \mathbb{R}^{n\times n}$得到的图滤波矩阵; 目标函数第一项表示投影后每个样本与其邻居线性组合之间的重建误差; 第二项用于将图滤波器与自表示矩阵联系起来,以利用数据高阶邻域信息进行自表示学习; 第三项为正则化项,强制投影矩阵 $W$ 为行稀疏以进行特征选择; $\alpha$ 和 $\lambda$ 为平衡第一项、第二项和第三项的正则化参数.

### 2.2 模型的求解

由于模型同时包含两个变量 $W$ 和 $Z$,直接求解非常困难,因此本文采用交替变量优化的方式解决该问题.

若固定 $Z$ 更新 $W$,则式(10)可转化为如下问题:

$$\begin{aligned}&\min \|\bar{X}W - Z\bar{X}W\|_{2,1} + \lambda\|W\|_{2,1},\\&\text{s.t. } W^TW = I.\end{aligned} \tag{11}$$

受 Nie 等[15]提出重加权方法的启发,式(11)可转换为

$$\begin{aligned}&\min \operatorname{tr}((\bar{X}W - Z\bar{X}W)^T D(\bar{X}W - Z\bar{X}W)) + \lambda\operatorname{tr}(W^TQW),\\&\text{s.t. } W^TW = I,\end{aligned} \tag{12}$$

其中 $D \in \mathbb{R}^{n\times n}$为对角矩阵且 $D_{ii} = \dfrac{1}{2\|\bar{X}_iW - Z_i\bar{X}W\|_2}$, $Q \in \mathbb{R}^{d\times d}$为对角矩阵且 $Q_{ii} = \dfrac{1}{2\sqrt{\|W_i\|_2^2}}$. 进一步模型可简化为

$$\begin{aligned}&\min \operatorname{tr}(W^T(P + \lambda Q)W)\\&\text{s.t. } W^TW = I,\end{aligned} \tag{13}$$

其中 $P = \bar{X}^T(I_n - Z)^T D(I_n - Z)\bar{X}$. 通过式(13)可轻松得到 $W$ 的最优解为$(P + \lambda Q)$的 $c$ 个最小特征值所对应的特征向量.

若固定 $W$ 更新 $Z$,则式(10)可转化为

$$\begin{aligned}&\min \|\bar{X}W - Z\bar{X}W\|_{2,1} + \alpha\|Z - A\|_F^2,\\&\text{s.t. } Z\mathbf{1}_n = \mathbf{1}_n,\quad Z \geqslant 0.\end{aligned} \tag{14}$$

根据重加权方法,式(14)可转化为

$$\begin{aligned}&\min \operatorname{tr}((\bar{X}W - Z\bar{X}W)^T D(\bar{X}W - Z\bar{X}W)) + \alpha\operatorname{tr}((Z - A)^T(Z - A)),\\&\text{s.t. } Z\mathbf{1}_n = \mathbf{1}_n,\quad Z \geqslant 0,\end{aligned} \tag{15}$$

其中 $D \in \mathbb{R}^{n\times n}$为对角矩阵且 $D_{ii} = \dfrac{1}{2\|\bar{X}_iW - Z_i\bar{X}W\|_2}$. 进一步式(15)可简化为

$$\begin{aligned}&\min \operatorname{tr}(Z^TDZC) + \alpha\operatorname{tr}(Z^TZ) - 2\operatorname{tr}(Z^TE),\\&\text{s.t. } Z\mathbf{1}_n = \mathbf{1}_n,\quad Z \geqslant 0,\end{aligned} \tag{16}$$

其中 $C = \bar{X}WW^T\bar{X}^T$, $E = D\bar{X}WW^T\bar{X}^T + \alpha A$. 显然式(16)是一个具有非负和线性约束的二次规划问题,很难直接求解,因此可借助 ADMM 算法解决该问题. 首先引入 Lagrange 乘子 $H = Z, \mu, \Sigma$ 得到以下问题:

$$\begin{aligned}&\min \operatorname{tr}(Z^TDHC) + \alpha\operatorname{tr}(Z^TH) - 2\operatorname{tr}(Z^TE) + \frac{\mu}{2}\left\|Z - H + \frac{1}{\mu}\Sigma\right\|_F^2,\\&\text{s.t. } Z\mathbf{1}_n = \mathbf{1}_n,\quad Z \geqslant 0,\quad H \geqslant 0,\end{aligned} \tag{17}$$

其中 $\mu$ 为正则系数,矩阵 $\Sigma \in \mathbb{R}^{n\times n}$用于刻画目标变量与辅助变量之间的差异. 由于式(17)包含多个变量,因此本文采用交替变量优化方法求解该问题.

1) 固定其他变量更新 $Z$,式(17)可转化为

$$\begin{aligned}&\min \|Z - M\|_F^2,\\&\text{s.t. } Z\mathbf{1}_n = \mathbf{1}_n,\quad Z \geqslant 0,\end{aligned} \tag{18}$$

其中 $M=H-\frac{1}{\mu}\boldsymbol{\Sigma}-\frac{1}{2}(DHC+\alpha H-2E)$. 该问题为具有基数约束的单纯性上欧氏投影问题，可通过已有的算法[16]有效解决.

2）固定其他变量更新 $H$，式(17)可转化为

$$\begin{aligned}&\min \|\boldsymbol{H}-\boldsymbol{N}\|_F^2,\\&\text{s.t. } \boldsymbol{H}\geqslant 0,\end{aligned} \tag{19}$$

其中 $N=Z+\frac{1}{\mu}\boldsymbol{\Sigma}-\frac{1}{\mu}(B^{T}ZC^{T}+\alpha Z)$，可得 $H$ 的解为

$$H_{ij}=\max\{N_{ij},0\}. \tag{20}$$

3）在更新变量 $Z$ 和 $H$ 后，需要更新 ADMM 算法参数 $\mu$ 和 $\boldsymbol{\Sigma}$:

$$\boldsymbol{\Sigma}=\boldsymbol{\Sigma}+\mu(\boldsymbol{Z}-\boldsymbol{H}), \tag{21}$$

$$\mu=p\mu, \tag{22}$$

其中 $p>1$ 是控制收敛速度的参数，$p$ 值越大，需要更少的迭代次数便可收敛，但可能降低最终目标函数的精度.

综上，可得:

算法 1　用 ADMM 算法求解式(16).

输入：自表示矩阵 $Z\in\mathbb{R}^{n\times n}$，对角矩阵 $D\in\mathbb{R}^{n\times n}$，矩阵 $C\in\mathbb{R}^{n\times n}$，矩阵 $E\in\mathbb{R}^{n\times n}$，参数 $\alpha$;

输出：自表示矩阵 $Z\in\mathbb{R}^{n\times n}$;

1）初始化 $u=1$，$p=1.01$，$\boldsymbol{\Sigma}=\boldsymbol{0}_{n\times n}$，$H=Z$;

2）while not do convergence

3）利用式(18)更新 $Z$;

4）利用式(20)更新 $H$;

5）利用式(21)更新 $\boldsymbol{\Sigma}$;

6）利用式(22)更新 $\mu$;

7）end while.

根据上述描述，本文提出的基于图滤波与自表示的无监督特征选择算法描述如下.

算法 2　基于图滤波与自表示的无监督特征选择算法.

输入：数据矩阵 $X$，簇个数 $c$，参数 $\alpha,\lambda$;

输出：特征选择矩阵 $W$，计算特征权重 $f_i=\|W_i\|_2(i=1,2,\cdots,d)$，并以降序排序，然后选取排名前 $m$ 个特征作为特征选择的结果;

1）随机初始化：投影矩阵 $W\in\mathbb{R}^{d\times c}$，自表示矩阵 $Z\in\mathbb{R}^{n\times n}$;

2）构造 $k$ 近邻图并利用式(4)计算图滤波器 $A\in\mathbb{R}^{n\times n}$;

3）while not do convergence

4）求解式(13)更新 $W$，$W$ 最优解为 $(A+\lambda Q)$ 的 $c$ 个最小特征值对应的 $c$ 个最小特征向量;

5）通过算法 1 更新 $Z$;

6）更新对角矩阵 $D$，其对角元素为 $D_{ii}=\frac{1}{2\|\bar{X}_iW-Z_i\bar{X}W\|_2}$;

7）更新对角矩阵 $Q$，其对角元素为 $Q_{ii}=\frac{1}{2\sqrt{\|W_i\|_2^2}}$;

8）判断收敛 $\left|\frac{F(t-1)-F(t)}{F(t)}\right|\leqslant 10^{-4}$，$F$ 为式(10)的目标函数;

9）end while.

# 3　实验结果分析

## 3.1　数据集

为衡量本文提出的基于图滤波与自表示的无监督特征选择算法的性能，本文将在 9 个基准数据集上进行实验，这些数据集都是特征选择算法常用的数据集，各数据集信息列于表 1.

表1 数据集信息

Table 1 Information of datasets

个

| 数据集 | 样本数 | 特征数 | 类别数 | 数据集 | 样本数 | 特征数 | 类别数 |
|---|---|---|---|---|---|---|---|
| GARBER | 66 | 4 553 | 4 | YALE | 165 | 1 024 | 15 |
| LUNG_DISCRETE | 73 | 325 | 7 | CSTR | 476 | 1 000 | 4 |
| KHAN | 83 | 1 069 | 4 | MADELON | 2 600 | 500 | 2 |
| LYMPHOMA | 96 | 4 026 | 9 | AR2600 | 2 600 | 2 200 | 100 |
| WARPAR10P | 130 | 2 400 | 10 | | | | |

### 3.2 实验对比方法及参数设置

为验证本文算法的有效性，将其与一个基线(ALLFEA)和7种具有代表性的无监督特征选择算法进行对比分析，对比算法包括 LapScore(Laplacian score)[17]，MCFS(multi-cluster unsupervised feature selection)[18]，SOGFS(structured optimal graph feature selection)[19]，LLEScore[20]，SRCFS(subspace randomization and collaboration feature selection)[21]，UDPFS(unsupervised discriminative projection for feature selection)[22]和BSFS(balanced spectral feature selection)[23]. 构造 $k$ 近邻图时设置 $k=5$. 对于本文涉及的两个平衡参数 $\alpha$ 和 $\lambda$，通过网格搜索在 $[10^{-3},10^{-2},\cdots,10^{3}]$ 内选定其值进行特征选择. 对其他比较算法，按照原文献中的原则调整参数，并且将所有算法中类簇数量设置为类的真实数量.

### 3.3 实验结果分析

对选择的特征，本文通过3个广泛使用的聚类性能度量评估 $k$ 均值聚类的性能，即准确率(ACC)、归一化互信息(NMI)和纯度(Purity). 为降低 $k$ 均值聚类受初始化的影响，随机初始化重复聚类20次并记录平均结果. 由于特征选择的最优数量未知，因此为更合理地评估本文无监督特征选择算法，本文记录了选择不同数量特征[10:10:100]的平均聚类性能，结果分别列于表2~表4.

表2 不同算法的ACC实验结果对比

Table 2 Comparison of ACC experimental results of different algorithms

| 数据集及平均值 | ALLFEA | LapScore | MCFS | SOGFS | LLEScore | SRCFS | UDPFS | BSFS | GFASR |
|---|---|---|---|---|---|---|---|---|---|
| GARBER | 59.09 | 58.91 | 55.71 | 59.04 | 54.84 | 58.67 | 65.02 | 68.19 | 71.79 |
| LUNG_DISCRETE | 63.42 | 54.93 | 64.73 | 64.84 | 58.29 | 56.08 | 64.91 | 65.86 | 70.14 |
| KHAN | 62.35 | 74.52 | 70.48 | 53.10 | 50.49 | 65.85 | 65.08 | 70.69 | 76.94 |
| LYMPHOMA | 53.39 | 50.27 | 52.99 | 53.86 | 44.78 | 50.14 | 48.92 | 48.53 | 55.77 |
| WARPAR10P | 24.38 | 29.84 | 22.76 | 25.43 | 30.56 | 28.42 | 34.73 | 31.05 | 36.00 |
| YALE | 38.94 | 39.12 | 40.63 | 43.63 | 31.74 | 34.35 | 41.10 | 34.28 | 43.88 |
| CSTR | 45.68 | 40.95 | 41.41 | 41.13 | 39.09 | 45.01 | 41.47 | 44.79 | 46.67 |
| MADELON | 52.48 | 51.92 | 50.83 | 56.03 | 52.04 | 51.76 | 54.04 | 58.63 | 61.21 |
| AR2600 | 13.21 | 10.99 | 13.05 | 16.10 | 17.03 | 12.22 | 12.60 | 13.21 | 18.25 |
| **平均值** | 45.88 | 45.72 | 45.84 | 45.91 | 42.10 | 44.72 | 47.54 | 48.36 | 53.40 |

表3 不同算法的NMI实验结果对比

Table 3 Comparison of NMI experimental results of different algorithms

| 数据集及平均值 | ALLFEA | LapScore | MCFS | SOGFS | LLEScore | SRCFS | UDPFS | BSFS | GFASR |
|---|---|---|---|---|---|---|---|---|---|
| GARBER | 15.85 | 33.93 | 32.87 | 22.08 | 19.14 | 37.98 | 30.40 | 47.70 | 48.08 |
| LUNG_DISCRETE | 60.11 | 51.25 | 61.00 | 60.61 | 53.93 | 52.78 | 61.30 | 61.89 | 66.42 |
| KHAN | 46.11 | 63.74 | 59.38 | 37.12 | 63.74 | 53.41 | 49.83 | 58.51 | 66.29 |
| LYMPHOMA | 56.01 | 55.69 | 55.87 | 58.71 | 48.39 | 55.47 | 49.22 | 47.74 | 59.25 |
| WARPAR10P | 20.36 | 29.94 | 18.57 | 22.33 | 31.22 | 28.38 | 34.32 | 27.03 | 37.36 |
| YALE | 44.36 | 44.52 | 46.30 | 50.71 | 37.97 | 40.76 | 47.33 | 40.11 | 50.57 |
| CSTR | 19.83 | 18.58 | 18.68 | 11.43 | 12.62 | 22.39 | 11.74 | 21.04 | 24.08 |
| MADELON | 0.56 | 0.41 | 0.17 | 1.08 | 0.22 | 0.37 | 0.96 | 2.37 | 3.71 |
| AR2600 | 41.36 | 36.54 | 40.58 | 43.76 | 45.47 | 39.05 | 39.41 | 40.56 | 48.33 |
| **平均值** | 33.84 | 37.18 | 37.05 | 34.20 | 34.74 | 36.73 | 36.06 | 38.55 | 44.90 |

表 4　不同算法的 Purity 实验结果对比

Table 4　Comparison of Purity experimental results of different algorithms

| 数据集及平均值 | ALLFEA | LapScore | MCFS | SOGFS | LLEScore | SRCFS | UDPFS | BSFS | GFASR |
|---|---|---|---|---|---|---|---|---|---|
| GARBER | 67.80 | 80.30 | 78.12 | 71.09 | 68.25 | 82.27 | 77.05 | 84.32 | 84.64 |
| LUNG_DISCRETE | 70.68 | 63.08 | 71.36 | 71.10 | 66.00 | 63.34 | 71.71 | 72.21 | 76.06 |
| KHAN | 65.18 | 77.03 | 74.63 | 55.65 | 53.87 | 69.16 | 71.15 | 73.78 | 78.98 |
| LYMPHOMA | 74.53 | 73.26 | 76.56 | 76.82 | 70.71 | 72.83 | 70.96 | 70.43 | 79.04 |
| WARPAR10P | 25.35 | 30.91 | 24.01 | 26.30 | 31.83 | 29.20 | 36.62 | 32.76 | 38.30 |
| YALE | 40.55 | 41.46 | 42.58 | 45.09 | 34.46 | 36.50 | 43.03 | 36.59 | 45.90 |
| CSTR | 48.87 | 47.17 | 47.44 | 43.80 | 43.23 | 50.39 | 43.71 | 50.81 | 52.71 |
| MADELON | 52.48 | 51.92 | 50.83 | 56.03 | 52.04 | 51.76 | 54.04 | 58.63 | 61.21 |
| AR2600 | 13.81 | 11.93 | 13.77 | 17.50 | 19.32 | 13.33 | 13.78 | 14.07 | 20.36 |
| 平均值 | 51.03 | 53.01 | 53.25 | 51.49 | 48.86 | 52.09 | 53.56 | 54.85 | 59.69 |

表 2~表 4 中最后一行显示了在 9 个基准数据集上所有算法的平均结果. 由表 2~表 4 可见：与使用所有特征的聚类相比，本文算法可以在少于 100 个特征的情况下获得类似甚至更优的结果；本文算法始终比其他无监督特征选择算法的性能更好. 本文算法在 ACC，NMI 和 Purity 指标上与第二最佳算法相比分别提高了 10.42%，16.47%，8.82%. 实验结果很好地说明了本文算法的有效性，不仅可以极大减少特征数量，减少计算负担，而且可以在一定程度上提高聚类性能.

图 1~图 3 为不同特征选择算法在每个数据集上对应于不同选择特征数量［10：10：100］的 ACC，

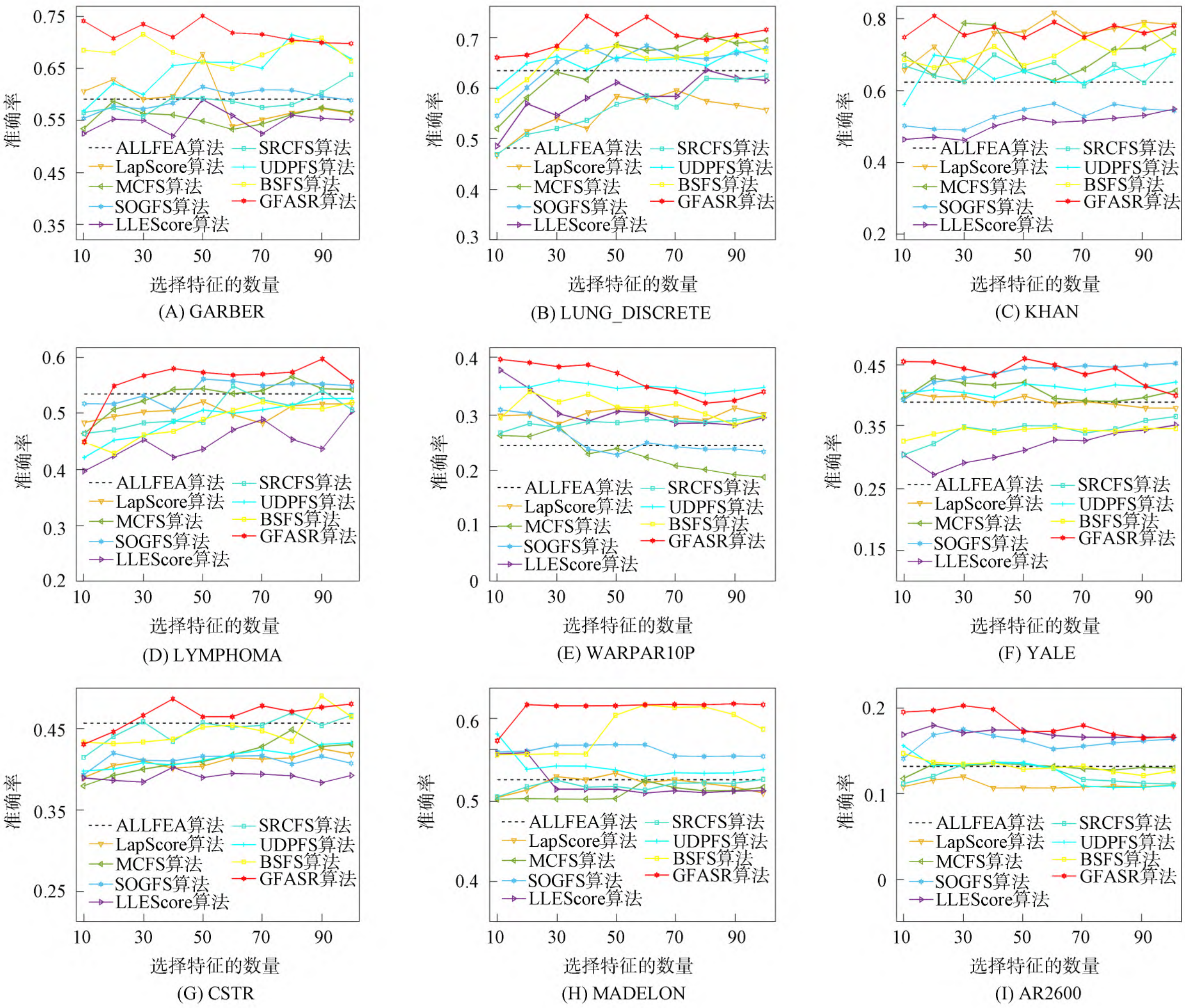

图 1　不同算法在不同数据集上特征选择的聚类准确率（ACC）

Fig.1　Clustering accuracy（ACC）of feature selection by different algorithms on different datasets

NMI 和 Purity 的实验结果.

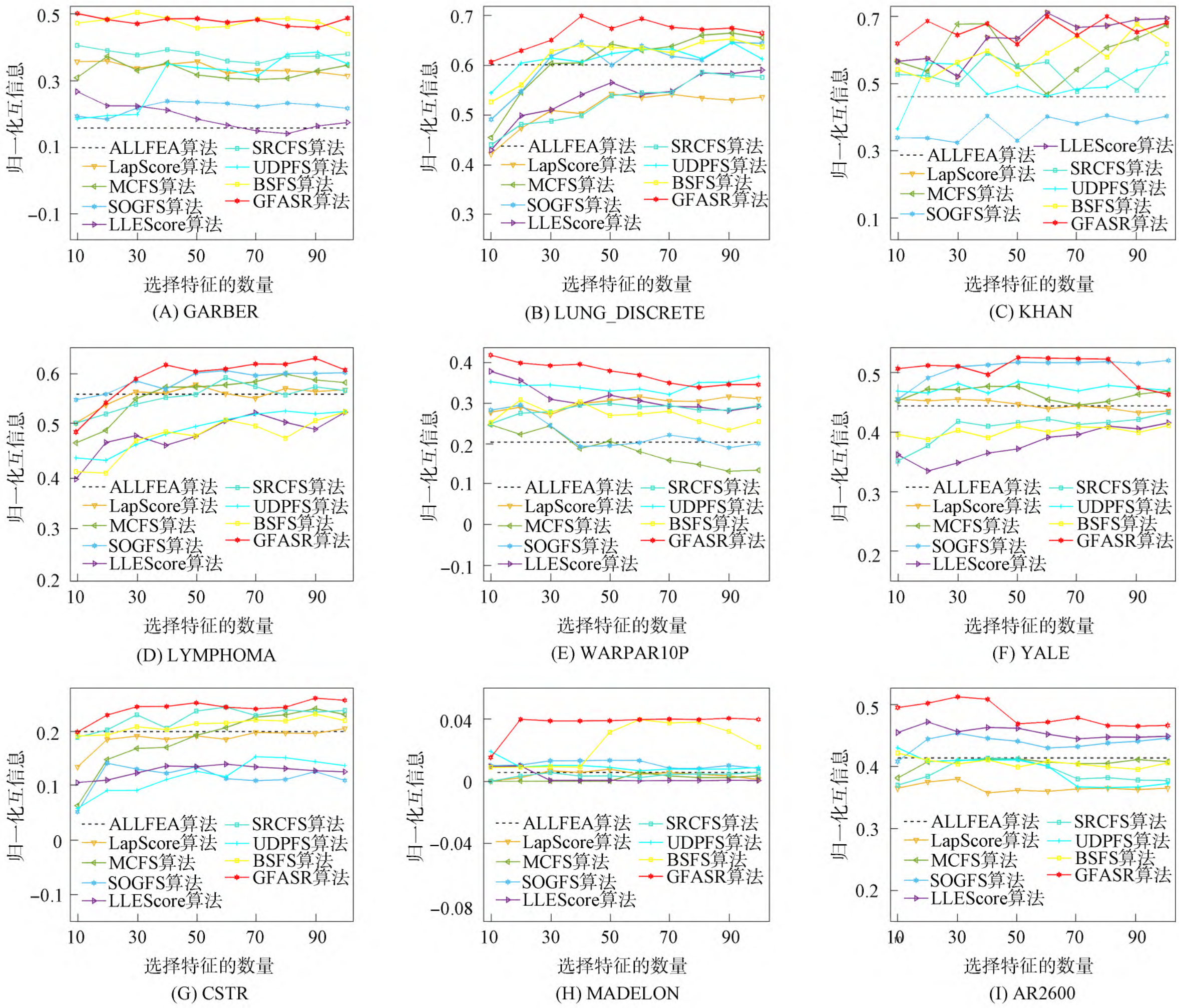

**图 2　不同算法在不同数据集上特征选择的归一化互信息(NMI)**

**Fig.2　Normalized mutual information (NMI) of feature selection by different algorithms on different datasets**

图中红线表示本文方法，黑色水平虚线表示使用所有特征的聚类结果. 由图 1~图 3 可见，在所有数据集上本文算法都优于 ALLFEA，表明本文算法不仅可以极大减少用于聚类的特征数量，而且可提高聚类性能. 虽然本文算法不能在每个特征选择数量上都超过对比算法，但在大多数数据集上都优于所对比的特征选择算法，表明本文提出对数据应用图滤波获得其平滑表示以及数据的高阶邻域信息有助于选择高质量鉴别特征.

本文参数敏感性实验在数据集 LUNG_DISCRETE 上进行可视化展示，如图 4 所示，其他数据集的结果类似. 由图 4 可见，本文算法在较宽的范围内性能相对稳定，因此，参数 $\alpha$ 和 $\lambda$ 的选择在实践中并不困难.

综上所述，针对现有方法未考虑数据的高阶邻域信息而不能完全捕捉数据内在结构的问题，本文提出了一种基于图滤波与自表示的无监督特征选择算法. 首先将图滤波器作用于原始数据获得其平滑表示，并设计一个正则化器利用隐藏在原始图中的高阶邻域信息，然后同时执行自表示学习与特征选择. 为增强重建误差项与特征选择项的鲁棒性和行稀疏性，用 $l_{1,2}$ 范数重建误差项和特征选择项. 在多个基准数据集上与多个先进的特征选择算法进行实验对比分析的结果表明，本文算法性能良好，验证了图滤波器作用于数据产生的平滑表示与原始图中隐藏的高阶邻域信息有助于特征选择任务.

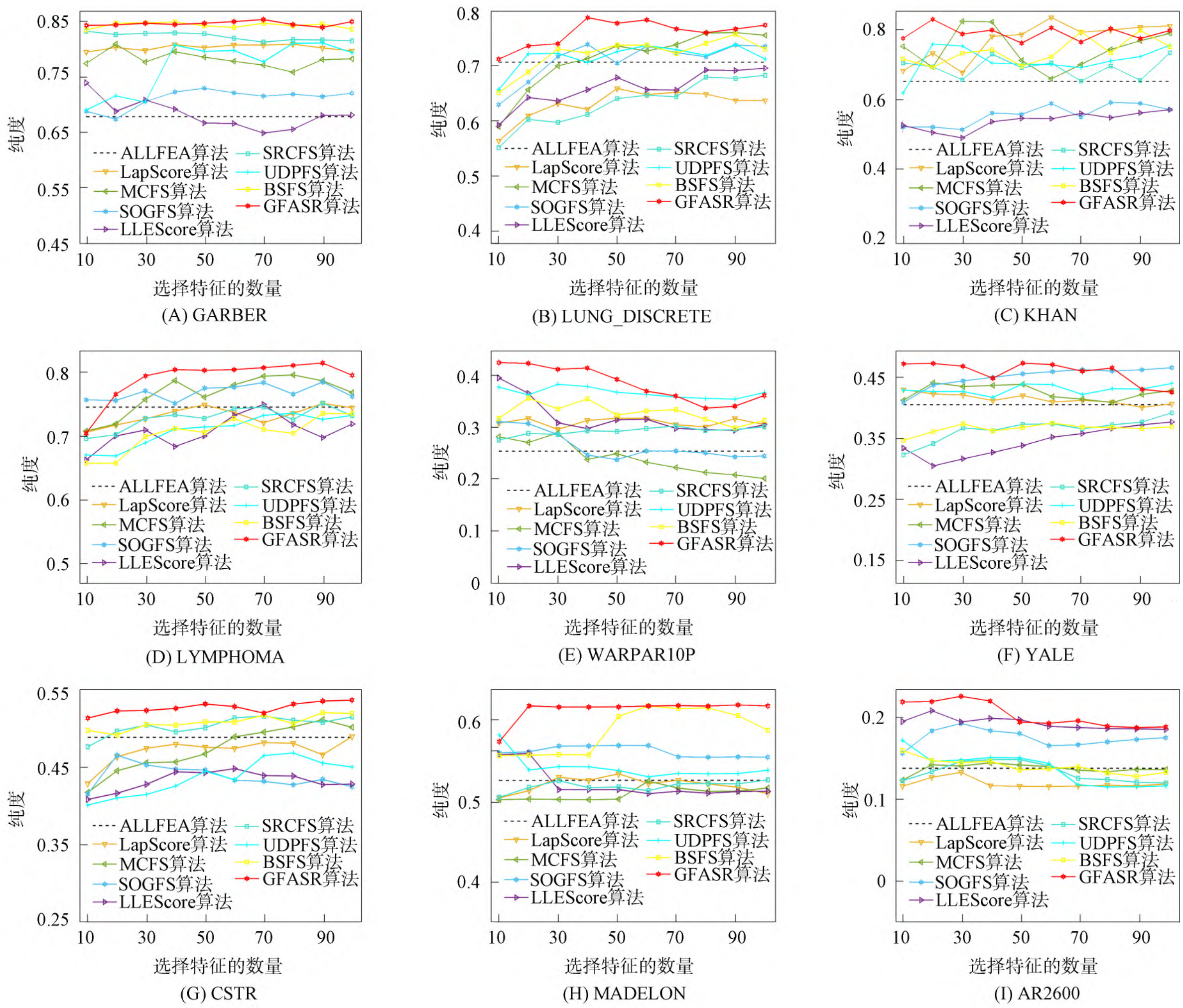

**图 3　不同算法在不同数据集上特征选择的聚类纯度(Purity)**

**Fig.3　Clustering purity (Purity) of feature selection by different algorithms on different datasets**

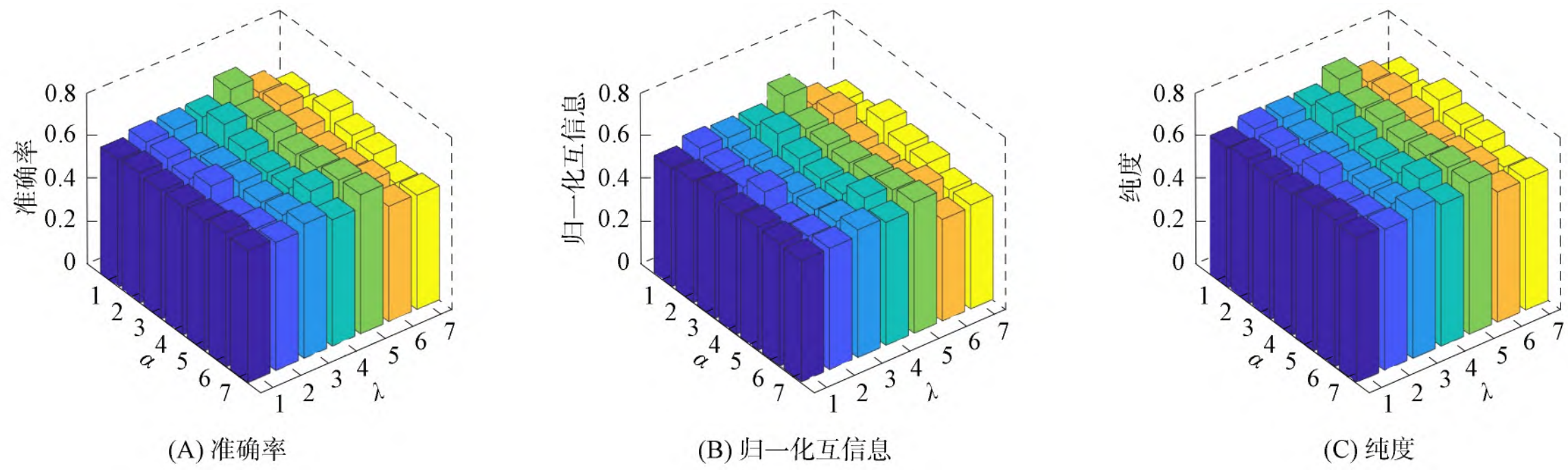

**图 4　在数据集 LUNG_DISCRETE 上参数选择对算法 2 的影响**

**Fig.4　Effect of parameters selection to algorithm 2 on LUNG_DISCRETE dataset**

## 参　考　文　献

[1]　YANG Y M, PEDERSEN J O. A Comparative Study on Feature Selection in Text Categorization [C] //Proceedings of the Fourteenth International Conference on Machine Learning. New York: ACM, 1997: 412-420.

[2]　RUI Y, HUANG T S, CHANG S F. Image Retrieval: Current Techniques, Promising Directions, and Open Issues [J]. Journal of Visual Communication and Image Representation, 1999, 10(1): 39-62.

[3]　SAEYS Y, INZA I, LARRANAGA P. A Review of Feature Selection Techniques in Bioinformatics [J]. Bioinformatics,

2007, 23(19): 2507-2517.

[4] CHEN J C, ZENG Y J, LI Y, et al. Unsupervised Feature Selection Based Extreme Learning Machine for Clustering [J]. Neurocomputing, 2020, 386: 198-207.

[5] MIAO J Y, YANG T J, SUN L J, et al. Graph Regularized Locally Linear Embedding for Unsupervised Feature Selection [J]. Pattern Recognition, 2022, 122: 108299-1-108299-13.

[6] CHEN H, NIE F P, WANG R, et al. Unsupervised Feature Selection with Flexible Optimal Graph [J]. IEEE Transactions on Neural Networks and Learning Systems, 2022, 35(2): 2014-2027.

[7] YUAN A H, YOU M B, HE D J, et al. Convex Non-negative Matrix Factorization with Adaptive Graph for Unsupervised Feature Selection [J]. IEEE Transactions on Cybernetics, 2022, 52(6): 5522-5534.

[8] YOU M B, YUAN A H, ZOU M, et al. Robust Unsupervised Feature Selection via Multi-group Adaptive Graph Representation [J]. IEEE Transactions on Knowledge and Data Engineering, 2021, 35(3): 3030-3044.

[9] ZHANG R, ZHANG Y X, LI X L. Unsupervised Feature Selection via Adaptive Graph Learning and Constraint [J]. IEEE Transactions on Neural Networks and Learning Systems, 2022, 33(3): 1355-1362.

[10] DU L, SHEN Y D. Unsupervised Feature Selection with Adaptive Structure Learning [C]//Proceedings of the 21th ACM SIGKDD International Conference on Knowledge Discovery and Data Mining. New York: ACM, 2015: 209-218.

[11] NIE F P, DONG X, TIAN L, et al. Unsupervised Feature Selection with Constrained $l_{2,0}$-Norm and Optimized Graph [J]. IEEE Transactions on Neural Networks and Learning Systems, 2022, 33(4): 1702-1713.

[12] ZHU P F, ZUO W M, ZHANG L, et al. Unsupervised Fature Selection by Regularized Self-representation [J]. Pattern Recognition, 2015, 48(2): 438-446.

[13] ZHAO Z, HE X F, CAI D, et al. Graph Regularized Feature Selection with Data Reconstruction [J]. IEEE Transactions on Knowledge and Data Engineering, 2015, 28(3): 689-700.

[14] LI J D, TANG J L, LIU H. Reconstruction-Based Unsupervised Feature Selection: An Embedded Approach [C]//Proceedings of the 26th International Joint Conference on Artificial Intelligence. New York: ACM, 2017: 2159-2165.

[15] NIE F P, HUANG H, CAI X, et al. Efficient and Robust Feature Selection via Joint $l_{2,1}$-Norms Minimization [C]//Advances in Neural Information Processing Systems. New York: ACM, 2010: 1813-1821.

[16] WANG W R, CARREIRA-PERPINÁN M A. Projection onto the Probability Simplex: An Efficient Algorithm with a Simple Proof, and an Application [EB/OL]. (2013-09-06) [2023-02-01]. https://arxiv.org/abs/1309.154.

[17] HE X F, CAI D, NIYOGI P. Laplacian Score for Feature Selection [C]//Advances in Neural Information Processing Systems. New York: ACM, 2005: 507-514.

[18] CAI D, ZHANG C Y, HE X F. Unsupervised Feature Selection for Multi-cluster Data [C]//Proceedings of the 16th ACM SIGKDD International Conference on Knowledge Discovery and Data Mining. New York: ACM, 2010: 333-342.

[19] NIE F P, ZHU W, LI X L. Unsupervised Feature Selection with Structured Graph Optimization [C]//Proceedings of the AAAI Conference on Artificial Intelligence. Palo Alto: AAAI, 2016: 1302-1308.

[20] YAO C, LIU Y F, JIANG B, et al. LLE Score: A New Filter-Based Unsupervised Feature Selection Method Based on Nonlinear Manifold Embedding and Its Application to Image Recognition [J]. IEEE Transactions on Image Processing, 2017, 26(11): 5257-5269.

[21] HUANG D, CAI X, WANG C D. Unsupervised Feature Selection with Multi-subspace Randomization and Collaboration [J]. Knowledge-Based Systems, 2019, 182: 104856-1-104856-15.

[22] WANG R, BIAN J T, NIE F P, et al. Unsupervised Discriminative Projection for Feature Selection [J]. IEEE Transactions on Knowledge and Data Engineering, 2022, 34(2): 942-953.

[23] ZHOU P, CHEN J Y, DU L, et al. Balanced Spectral Feature Selection [J]. IEEE Transactions on Cybernetics, 2023, 53(7): 4232-4244.